\documentclass[runningheads]{llncs}
\usepackage[T1]{fontenc}
\usepackage{graphicx}
\usepackage{wrapfig}  
\usepackage{comment}
\usepackage{multirow}
\usepackage{booktabs}
\usepackage{siunitx}
\usepackage{soul}

\begin{document}
\title{Robustness and Diagnostic Performance of Super-Resolution Fetal Brain MRI}

%\title{Evaluating Super-Resolution Methods for Fetal Brain MRI: Volumetry, Success Rate, and Classification}
%
\titlerunning{Robustness and Diagnostic Performance of SRR Methods}
% If the paper title is too long for the running head, you can set
% an abbreviated paper title here
%
\author{Ema Masterl\inst{1,\dagger}\orcidID{0009-0007-4115-4304} \and
Tina Vesnaver Vipotnik\inst{2} \and\\
Žiga Špiclin\inst{3}}
\authorrunning{E. Masterl et al.}
% First names are abbreviated in the running head.
% If there are more than two authors, 'et al.' is used.
%
\institute{Faculty of Medicine, University of Ljubljana, %Vrazov trg 2, SI-1000 
Ljubljana, Slovenia \and
University Medical Centre Ljubljana, %Zaloška cesta 7, SI-1000 
Ljubljana, Slovenia \and
Faculty of Electrical Engineering, University of Ljubljana, %Tržaška cesta 25, SI-1000 
Ljubljana, Slovenia\\
$^\dagger$\textit{Corresponding author:} \email{ema.masterl@mf.uni-lj.si}}

% \author{* * * * * * * \inst{1} \and
% * * * * * * * * * * * *\inst{2} \and\\
% * * * * * * * \inst{3,\dagger}\orcidID{****-****-****-****}}
% %
% \authorrunning{* * * * * * et al.}
% % First names are abbreviated in the running head.
% % If there are more than two authors, 'et al.' is used.
% %
% \institute{* * * * * * * * * * * * * * * * * * * * * * %Vrazov trg 2, SI-1000 
% * * * * * * * * * * \and
% * * * * * * * * * * * * * * * * * *
% * * * * * * * * * * \and
% * * * * * * * * * * * * * * * * * * * * * * * * * * * * * *
% * * * * * * * * * *\\
% $^\dagger$\textit{Corresponding author:} \email{anonymous@anonymous.org}}

%
\maketitle              % typeset the header of the contribution
\begin{abstract}
Fetal brain MRI relies on rapid multi-view 2D slice acquisitions to reduce motion artifacts caused by fetal movement. However, these stacks are typically low resolution, may suffer from motion corruption, and do not adequately capture 3D anatomy. Super-resolution reconstruction (SRR) methods aim to address these limitations by combining slice-to-volume registration and super-resolution techniques to generate high-resolution (HR) 3D volumes. While several SRR methods have been proposed, their comparative performance—particularly in pathological cases—and their influence on downstream volumetric analysis and diagnostic tasks remain underexplored. In this study, we applied three state-of-the-art SRR methods—NiftyMIC, SVRTK, and NeSVoR—to 140 fetal brain MRI scans, including both healthy controls (HC) and pathological cases (PC) with ventriculomegaly (VM). Each HR reconstruction was segmented using the BoUNTi algorithm to extract volumes of nine principal brain structures. We evaluated visual quality, SRR success rates, volumetric measurement agreement, and diagnostic classification performance. NeSVoR demonstrated the highest and most consistent reconstruction success rate (>90\%) across both HC and PC groups. Although significant differences in volumetric estimates were observed between SRR methods, classification performance for VM was not affected by the choice of SRR method. These findings highlight NeSVoR’s robustness and the resilience of diagnostic performance despite SRR-induced volumetric variability.

\end{abstract}

\section{Introduction}
Fetal MRI acquisitions employ rapid multi-view 2D slice stacks to mitigate the effects of continuous fetal movement. While the acquisitions provide high tissue contrast ideal for studying brain development and anomaly detection, the multi-view slice stacks are low resolution, may contain motion-corrupted slices and generally do not visualize the 3D anatomy. To address these challenges, a range of super‐resolution reconstruction (SRR) methods has been introduced~\cite{ebner2020automated,kuklisova2012reconstruction,uus2024scanner,10015091,tourbier_automated_2017} that combine iterative slice-to-volume registration with super‐resolution to generate a single 3D high-resolution (HR) image. In context of fetal brain imaging, the SRR methods also involve a spatial co-registration of the 3D HR image to gestational age (GA) matched brain atlas image. However, the performance and impact of particular SRR method on the performance of obtained 3D HR image in downstream tasks, like brain structure segmentation, biometry and diagnostics, especially in pathological cases, requires further study.

Most previous studies comparatively assessed in context of fetal brain imaging two or more of the four widely used SRR methods: NiftyMIC~\cite{ebner2020automated}, SVRTK~\cite{kuklisova2012reconstruction,uus2023combined,uus2024scanner}, NeSVoR~\cite{10015091} and MIALSRTK~\cite{tourbier_automated_2017}.
For instance, Sanchez et al.~\cite{sanchez2024assessing} assessed the first three on 84 fetuses ranging from 20-36 weeks of GA and established that NeSVoR demonstrated the most consistent performance, with minimal bias field and overall image reliability. When assessing volumetric measurement biases between these methods, these were found small and systematic at 2.8\%~\cite{sanchez2024biometry}. While evaluating the three aforementioned SRR pipelines -- NiftyMIC, SVRTK and MIALSRTK -- on 17 healthy fetuses with GA of 20–21 weeks, Ciceri et al.~\cite{ciceri2023geometric} reported a rather low success rate of SVRTK of 44\%. Xu et al.~\cite{xu2023nesvor}, while comparing NeSVoR with SVRTK reconstructions based on automated quality assessment metrics and signal-to-noise ratio %, using 3–10 motion-corrupted stacks per subject 
across 20 fetuses in the 21–32 week GA range, the NeSVoR achieved superior image quality. In qualitative comparative assessment of NiftyMIC, SVRTK and MIALSRTK, Uus et al. \cite{uus2023retrospective} concluded that  image quality was comparable. However, studies included rather small sample sizes, rendering estimates of success rate unreliable and 
did not include pathological cases that could have adverse impact. The impact on volumetry and its diagnostic performance also remains unclear.

\begin{figure}
    \centering
    \includegraphics[width=1.0\linewidth]{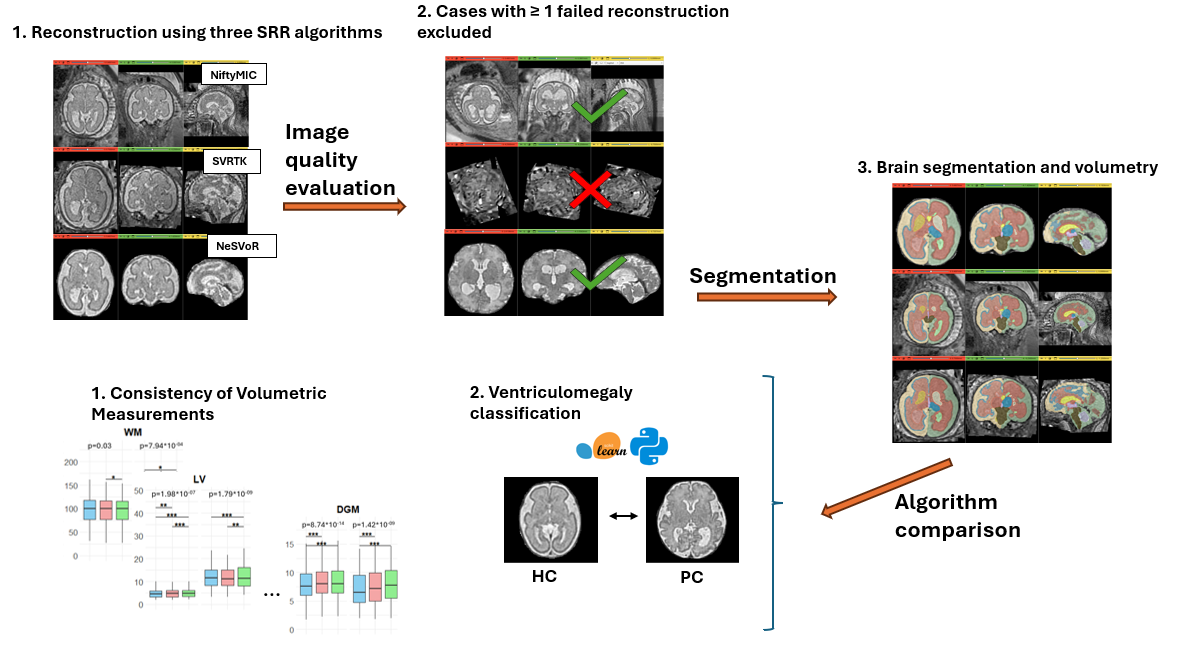}
    \caption{Evaluation pipeline comparing the SRR methods, encompassing reconstruction-quality assessment, volumetric analysis and diagnostic-performance testing.}
    \label{pipeline}
\end{figure}

In this study, we applied NiftyMIC \cite{ebner2020automated}, SVRTK \cite{kuklisova2012reconstruction,uus2023combined,uus2024scanner} and NeSVoR \cite{10015091} to 
140 fetal brain MRI acquisitions, % TODO: verify, since Data section says 140
comprising both healthy and pathological cases (HC and PCs, resp.) with ventriculomegaly (VM).  Following reconstruction, each 3D HR image was segmented using the BoUNTi method \cite{uus2023bounti} to extract 9 brain structures and corresponding volumes. Besides visual quality and SRR success rate assessment, we evaluated the agreement between volumetric measurements with respect to SRR method and their diagnostic performance for HC versus PC classification, using rigorous statistical analyses.  Study design is visualised in Figure \ref{pipeline}.

\section{Materials and Methods}

\subsection{Dataset}

A total of  140 fetal MRI acquisitions with GA from 21 to 37 weeks were retrospectively collected for this study. Inclusion criteria were: (1) only singletone pregnancies, (2) absence of excess fetal motion or artifacts, and (3) image stacks captured whole fetal brain. The cohort comprised fetuses with normally developing brain and those diagnosed with VM. Diagnosis of VM was re-confirmed by a neuroradiologist ($>$ 15 years of experience in reading fetal MRI) based on measuring the diameter, using linear measurement tool in OsiriX 12.0, the lateral ventricles in axial transventricular plane, and following the recommendations of the Society for Maternal-Fetal Medicine\cite{fox_mild_2018}, using a cut-off threshold of $\geq$ 10 mm for VM diagnosis. Stratification into mild/moderate/severe VM showed high imbalance towards mild cases, hence it was not used for the purposes of this study.

All imaging was performed at the University Medical Center Ljubljana on a 1.5 T clinical MRI scanner (Siemens Aera, Siemens Healthineers, Erlangen, Germany) using a T2-weighted Half-Fourier Acquisition Single-shot Turbo spin-Echo (HASTE) sequence. Each case included at least three orthogonal image stacks encompassing the entire fetal brain. In cases where either maternal or fetal motion was observed on a given stack, the acquisition of that stack was repeated to ensure adequate image quality. Acquisition parameters were set to achieve high in-plane resolution (0.625$\times$0.625 mm) with a slice thickness of 3 mm, resulting in image matrices of 512$\times$320 pixels and with 31--35 slices per stack. All data were anonymized prior to analysis.

The study protocol received approval from the Institutional Review Board (IRB; approval no.: 0120-56/2022/3). The IRB waived collection of informed consents for this retrospective, secondary data analysis study.  

\subsection{Super-Resolution Reconstruction}
Super‐resolution reconstruction was carried out using three established methods: NiftyMIC~\cite{ebner2020automated}, SVRTK~\cite{kuklisova2012reconstruction,uus2023combined,uus2024scanner} and NeSVoR~\cite{10015091}. Each SRR image was reconstructed at 0.5 mm isotropic resolution. 
For the NiftyMIC pipeline, each input stack was first skull-stripped to generate a brain mask; any masks deemed suboptimal were manually refined prior to reconstruction. 

During NiftyMIC processing, we also modified the default outlier slice rejection thresholds (\texttt{threshold}=0.1; \texttt{threshold\_first}=0.5). The SVRTK and NeSVoR methods were executed using default parameter settings.

All SRR images underwent visual quality control (VQC) in 3D Slicer\footnote{3D Slicer: \url{https://www.slicer.org/}} by a rater with over two years of fetal MRI reading experience. A successful VQC required complete whole‐brain coverage and absence of intraparenchymal signal dropouts (gaps) or other major artifacts within the brain parenchyma (blurry/distorted anatomy). Cases for which all three SRR methods resulted in a successful VQC were retained for subsequent analyses. Success rates were recorded.

\textbf{Statistical analysis.} Within the group of HC and PC cases, we used Fisher’s exact test to evaluate whether the proportion of correctly classified cases differed significantly between SRR methods. The Benjamini–Hochberg procedure was applied to control the false discovery rate. 

\subsection{Brain Segmentation and Volumetry}
Volumetric segmentation of the principal brain structures was carried out using the pre-trained BoUNTi algorithm~\cite{uus2023bounti}. Nine volumetric measurements were extracted for each case and each of three SRR images, corresponding to the major anatomical regions of interest: (1) extra-cerebral spinal fluid (ECSF), (2) grey matter/cerebral cortex (GM), (3) white cerebral matter (WM), (4) lateral ventricles (LV), (5) right LV (R\_LV), (6) left LV (L\_LV), (7) deep grey matter (DGM), 
(8) cerebellum (CRB), and (9) brainstem (BS). Volumetric measurements were recorded in cubic millimeters (mm$^3$).

\textbf{Statistical analysis.} To assess the agreement of volumetric measurements of principal brain structures across the three SRR methods, a non-parametric statistical analysis was conducted due to the non-normal distribution of the data. The Friedman test was employed for univariate repeated measures analysis to detect overall differences in volumetric estimates among the SRR methods for each brain structure. Analyses were performed separately for HCs and PCs to account for potential group-specific variability. Holm’s method was applied to adjust p-values for multiple comparisons, ensuring control of the family-wise error rate across the set of statistical tests.

In instances where the Friedman test indicated statistically significant differences (p < 0.05), post-hoc pairwise comparisons between SRR methods were performed using the Wilcoxon signed-rank test, with Holm correction for multiple comparisons. 

\subsection{Diagnosis Classification}
To evaluate the diagnostic performance of SRR fetal brain images, we performed a classification of HCs versus PCs using the volumetric measurements. % extracted from nine principal brain regions. 
For each SRR method (NiftyMIC, SVRTK, NeSVoR), a separate classification model was trained and evaluated using a 50:50\% case-wise stratified train:test split. The classification models were developed using the \texttt{auto-sklearn}~\cite{feurer-neurips15a} automated machine learning toolbox, which selected and optimized the best-performing model based on the training set volumetry corresponding to each SRR method.

The following key parameters were configured: \texttt{time\_left\_for\_this\_task} was set to 3600 seconds (1 hour)% to allow sufficient time for model exploration and ensemble construction%
, while \texttt{per\_run\_time\_limit} was set to 300 seconds to restrict the time allocated to individual model evaluations. The \texttt{ensemble\_size} and \texttt{ensemble\_nbest} parameters were at default values of 50, enabling \texttt{auto-sklearn} to construct an ensemble from the top 50 models based on validation performance. The \texttt{resampling\_strategy} was set to \texttt{holdout} to match the fixed 50:50 train\:test split used in our study design. All other parameters %, including preprocessing strategies and model types considered, 
were kept at their default settings. 

When best overall classification model was identified, it was fixed and the training repeated. This setup ensured automated model training and unbiased results across experiments runs.

% a standardized yet robust approach to automated model training and selection across the three SRR methods.

\textbf{Statistical analysis.} The trained models were evaluated on all test sets (across volumetry obtained with different SRR methods), and performance was assessed using area under the receiver operating characteristic curve (AUC), sensitivity, and specificity metrics. The optimal operating point on each ROC curve was identified using Youden’s index (maximizing sum of sensitivity and specificity minus one).

To compare diagnostic performance across SRR methods, we conducted pairwise statistical testing: the DeLong test was applied to compare AUCs, while McNemar’s test was used to assess differences in sensitivity and specificity. This framework enabled a systematic evaluation of the impact of SRR method choice on classification performance in distinguishing HC from PC based on quantitative brain volumetry.

\section{Experiments and Results}

\subsection{SRR Success Rates}

\begin{figure}[!ht]
    \centering
    \includegraphics[width=1.0\linewidth]{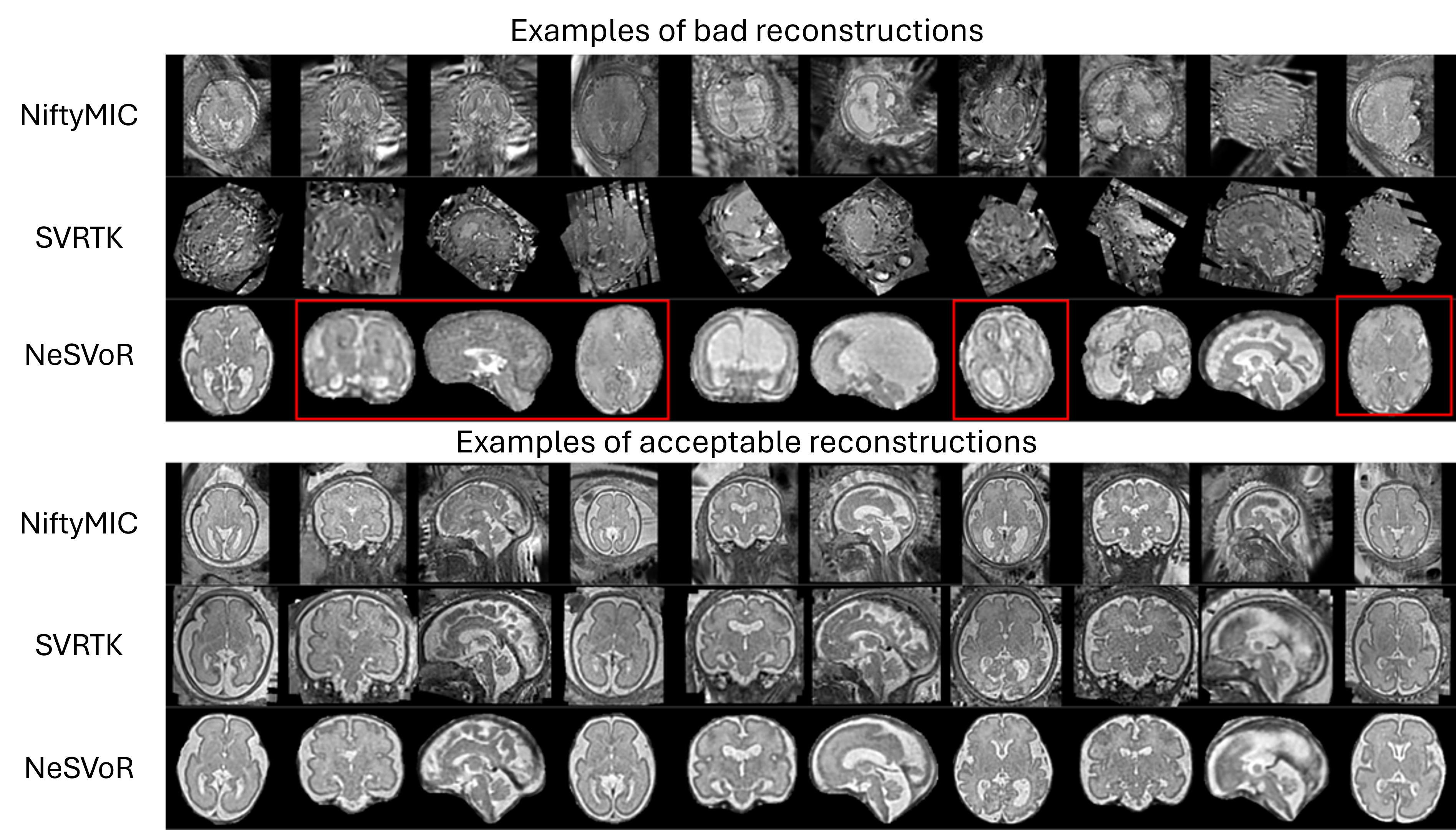}
    \caption{Discarded (\textit{top}) and successful (\textit{bottom}) cases in \textit{columns} with corresponding reconstructions for NiftyMIC, SVRTK, NeSVoR (\textit{1st, 2nd, 3rd} row, resp.). Some cases \textit{on top} with NeSVoR method were successful, those with \textit{red outline} are failed ones.}
    \label{reconstruction_example}
\end{figure}

Examples of reconstructions not passing and passing VQC are shown in Figure \ref{reconstruction_example}. Table \ref{rec_successfulness} shows the amount of successful reconstructions per each method. At the end, 95 of 140 cases were used for analysis, i.e. 44/50 HC (88.00\%) and 51/90 PC (56.67\%). The NeSVoR method had highest overall success rate, while in general the performance of all methods was lower for PC, with the most noticable drop when using the SVRTK method. 

Fisher’s exact tests revealed no significant differences in sucess rate between methods for HCs (p = 0.345). In contrast, the differences were statistically significant for the PCs (p = 0.00028). Post-hoc comparisons showed significant differences between SVRTK and NeSVoR (p = 0.00069) as well as between SVRTK and NiftyMIC (p = 0.013), while no significant difference was observed between NiftyMIC and NeSVoR (p = 0.372).

\begin{table}
\centering
\caption{Success rates of SRR methods, overall and stratified by HCs and PCs.}
\begin{tabular}{||c | c c c||} 
 \hline
  Group & NiftyMIC & SVRTK & NeSVoR \\ [0.5ex] 
 \hline\hline
 HC & 46 (92.0\%) & 45 (90.0\%) & 49 (98.0\%) \\ 
 \hline
 PC & 76 (84.4\%) & 60 (66.7\%) & 81 (90.0\%) \\
 \hline
  Overall & 122 (87.1\%) & 105 (75.0\%) & 130 (92.8\%) \\
 \hline
\end{tabular}
\label{rec_successfulness}
\end{table}

%%%%%%%%%%%%%%%%%%%%%%%%%%%%%%%%%%%%%%%%%%%%%%%%%%%%%%%%%%%%%%%%%
\subsection{Consistency of Volumetric Measurements}
\begin{figure}[!ht]
\centering
    \includegraphics[scale=0.6]{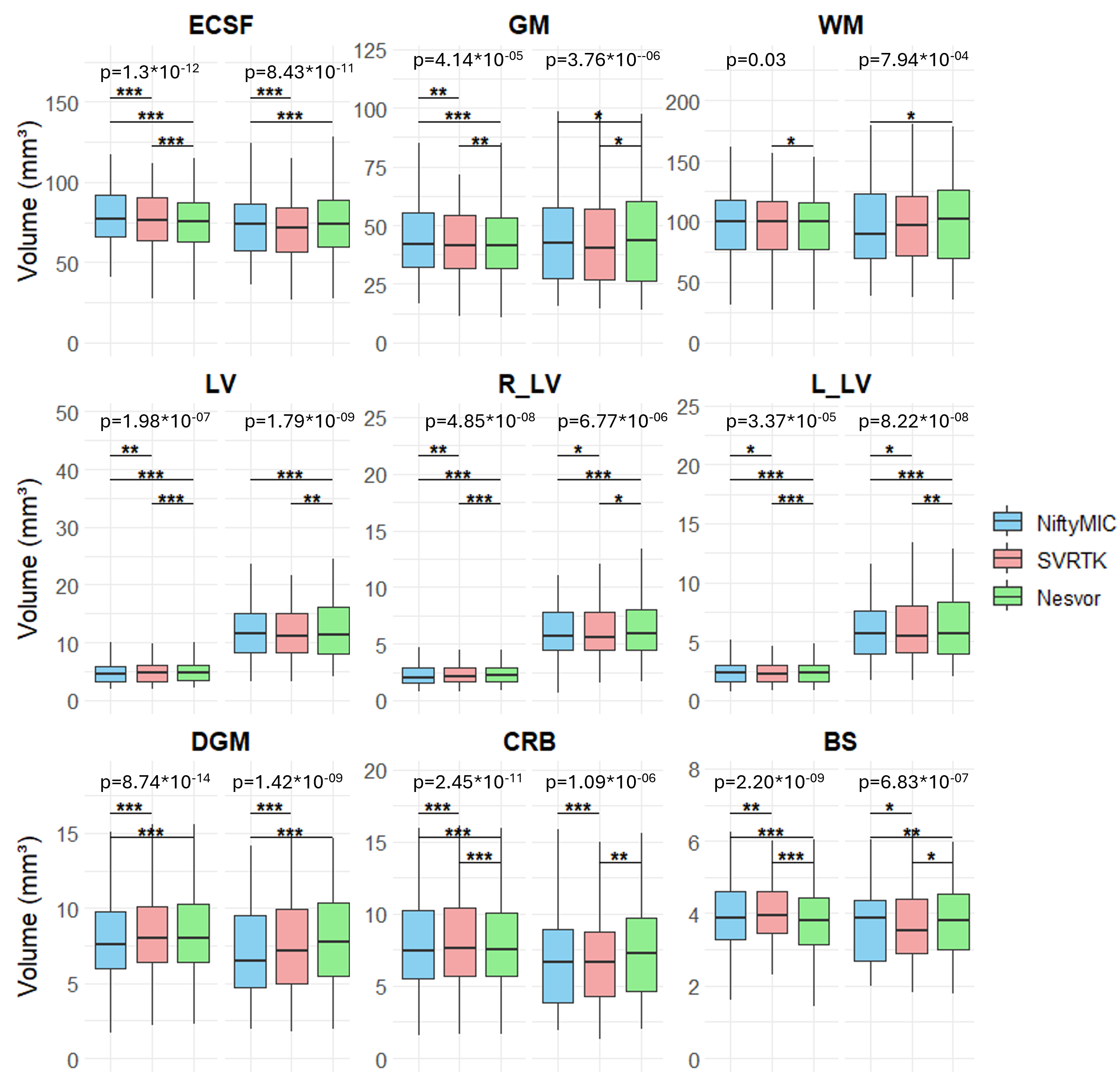}
    \caption{Box-whisker plots of 9 volumetric measurements for HC (\textit{left}) and PC (\textit{right}) groups. The p-values \textit{on top of each plot} are for Friedman test, while for post-hoc pairwise test they are indicated with \textit{bars} as *p<0.05, **p<0.01 and  ***p<0.001.}
    \label{boxplot}
\end{figure}

Volumetric measurements were analysed for 44 HCs and 51 PCs, in which all SRR methods were successful according to the VQC. Figure \ref{boxplot} shows the distribution of volumetric measurements for all SRR methods. The differences between HC and PC groups are most evident in LV, R\_LV and L\_LV volumes. Statistically significant differences were observed for each volumetric measurement when comparing the three SRR methods. The weakest significance was found in WM, where differences were the least pronounced. Lower p-values were generally observed in the HC group, with the strongest significance in DGM, ECSF, and CRB. In the PC group, the most prominent differences were found in ECSF, DGM, and LV.

%posthoc
Post hoc analysis revealed that volume differences were statistically significant for a greater number of structures in the HC group, with six volumes (ECSF, LV, R\_LV, L\_LV, CRB, and BS) showing significant differences across all three pairwise comparisons. In PC only the LV, R\_LV, and L\_LV showed consistent differences across all SRR methods, while the remaining structures exhibited differences in only a subset of pairwise comparisons. Notably, for ventricular volumes, NeSVoR consistently yielded most significant differences.

\subsection{Ventriculomegaly Classification}

Linear discriminant analysis was identified as the optimal classifier based on \texttt{auto-sklearn}. The 3D HR images as obtained by the three SRR methods, and upon based brain volumetry, deliver virtually identical diagnostic performance as shown in Table \ref{tab:model_validation_results}; namely, the AUC values clustered tightly in range 0.94--0.95 across every model $\times$ test-set pairing; sensitivity was constant at 0.96 on NiftyMIC- and SVRTK-derived test sets and only marginally lower on NeSVoR data (0.88 for NiftyMIC/SVRTK models, 0.92 for the NeSVoR model); specificity was lowest for the in-domain NiftyMIC and SVRTK evaluations (0.86) but rised to range 0.90--0.95 in cross-domain settings, keeping overall discrimination unchanged. DeLong's and McNemar's tests  both yielded p > 0.5, confirming that none of the reconstruction-specific models was statistically superior for  HC-versus-VM classification.

\begin{table}[ht]
\centering
\caption{Classification performance between HC and PC cases using \texttt{auto-sklearn}.}
\begin{tabular}{||l|c|cc|c|cc|c|cc||} 
 \hline
 SRR model / Test data
   & \multicolumn{3}{c|}{NiftyMIC} 
   & \multicolumn{3}{c|}{SVRTK} 
   & \multicolumn{3}{c||}{NeSVoR} \\ [0.5ex]
 \cline{2-10}
   & AUC & Sens & Spec & AUC & Sens & Spec & AUC & Sens & Spec \\
 \hline\hline
 NiftyMIC & 0.94 & 0.96 & 0.86 & 0.95 & 0.96 &  0.90 & 0.95 & 0.88 & 0.95 \\ 
 \hline
 SVRTK    & 0.94 & 0.96 & 0.86 & 0.95 & 0.96 &  0.90 & 0.95 & 0.88 & 0.95 \\  
 \hline
 NeSVoR   & 0.94 & 0.96 & 0.86 & 0.95 & 0.96 &  0.90 & 0.94 & 0.92 & 0.90 \\  
 \hline
 \multicolumn{10}{l}{\scriptsize Sens--Sensitivity; Spec--Specificity}
\end{tabular}
\label{tab:model_validation_results}
\end{table}

\section{Discussion}
This study evaluated the influence of three SRR methods — NiftyMIC, SVRTK and NeSVoR — on reconstruction quality, volumetric quantification of nine fetal brain volumes, and downstream  classification of VM.

All three SRR pipelines demonstrated high reconstruction success rates in HC. Fisher’s exact test confirmed that the success rates did not differ significantly. Reconstruction quality, however, declined for the PC. The lower success rate observed for SVRTK is consistent with Ciceri et al. \cite{ciceri2023geometric}, who reported unsuccessful reconstructions in 44 \% of the cases when using the SVRTK method. The highest success rate for both HC and PC was achieved by the NeSVoR method.

The non-parametric Friedman test demonstrated statistically significant inter-method differences in the volumetric estimates of all nine brain structures for both HCs and PCs. Post-hoc tests showed broader inter-method divergence in HC — significant for six structures—whereas in PC consistent differences were limited to ventricular volumes. This underscores the need to evaluate classification performance for additional fetal brain anomalies beyond ventriculomegaly.

Comparative evaluation of volumetric measurements across SRR methods, in the absence of a reliable ground truth, may be insufficient to assess their diagnostic performance. While systematic differences in volumetric estimates can be quantified, their practical relevance remains unclear without linking them to downstream tasks. For example, Sanchez et al.~\cite{sanchez2024biometry} reported small but consistent volumetric biases (~2.8\%) among the NiftyMIC, SVRTK and NeSVoR methods, as also evident from our analysis. However, such evaluations, based solely on inter-method discrepancies, do not necessarily inform the diagnostic performance of the measurements. 

In our study, despite observing statistically significant and systematic volumetric differences across SRR methods, the HC/VM classification performance remained unaffected. This finding underscores the need for evaluation protocols that extend beyond volumetric agreement and incorporate task-specific performance metrics. Statistical comparison of the classifiers trained on volumes reconstructed with NiftyMIC, SVRTK and NeSVoR methods revealed no significant differences in performance. Accordingly, for the binary task of distinguishing HCs from VM cases, all three SRR methods provided equivalent diagnostic performance. Whether this concordance persists when VM is stratified by severity (mild, moderate, severe) or when a different intracranial pathology is analysed remains undetermined. Ultimately, thoroughly assessing diagnostic performance requires validating whether method-dependent variability impacts decision-making outcomes, rather than focusing exclusively on measurement consistency.

An important consideration for strengthening the evaluation lies in the inclusion of additional quality control (QC) metrics, particularly segmentation accuracy. While intuitively appealing, evaluating segmentation accuracy is non-trivial in this context. Establishing a reference segmentation requires manual annotations, which are inherently subjective and labor-intensive, especially in fetal MRI where anatomical boundaries can be ambiguous. More critically, manual segmentations are typically performed on reconstructed images, making them inherently biased toward the SRR method used to generate the images and upon based reference segmentations. Any comparison of segmentation accuracy across SRR methods would thus conflate true segmentation performance with differences in image quality and reconstruction artifacts, undermining the validity of such  evaluation. Moreover, reconstruction errors—such as misalignment or motion artifacts—can propagate into the segmentations, further distorting the assessment. Given these challenges, while segmentation accuracy remains a desirable QC metric, its implementation in this setting would require careful protocol design to minimize method-specific biases, and may ultimately offer limited additional insight compared to task-based evaluations such as diagnostic classification.

The default parameter settings in NiftyMIC method generally results in a high rejection rate of input slices, which substantially increases the number of unsuccessful reconstructions. Similar observations were reported by Sanchez et al.~\cite{sanchez2024assessing}. To address this issue, parameter adjustment was performed in order to reduce the number of rejected slices and improve the overall reconstruction success rate. Conversely, the automated SVRTK method does not permit manual adjustment of key reconstruction parameters, which prevents empirical tuning based on systematic evaluation. Consequently, it is not possible to identify and apply a globally optimal parameter configuration that could improve reconstruction success rates across the dataset. This introduces a potential bias in the comparison between methods.

In conclusion, although the SRR methods yield virtually identical diagnostic performance for VM classification, NeSVoR attains the highest reconstruction success rate, compared to the NiftyMIC and SVRTK. In routine clinical examinations -- where fetal motion often compromises acquisition quality -- this superior robustness maximizes the likelihood of generating  usable high-resolution volume reconstructions without compromising diagnostic performance.

\section{Acknowledgements}
This study was supported by the Slovenian Research Agency (Core Research Grant No. P2-0232 and Research Grants No. and J2-3059).

%\FloatBarrier
% ---- Bibliography ----
\bibliographystyle{splncs04}
\bibliography{main}

\begin{thebibliography}{10}
\providecommand{\url}[1]{\texttt{#1}}
\providecommand{\urlprefix}{URL }
\providecommand{\doi}[1]{https://doi.org/#1}

\bibitem{ciceri2023geometric}
Ciceri, T., Squarcina, L., Pigoni, A., Ferro, A., Montano, F., Bertoldo, A., Persico, N., Boito, S., Triulzi, F.M., Conte, G., et~al.: Geometric reliability of super-resolution reconstructed images from clinical fetal mri in the second trimester. Neuroinformatics  \textbf{21}(3),  549--563 (2023)

\bibitem{ebner2020automated}
Ebner, M., Wang, G., Li, W., Aertsen, M., Patel, P.A., Aughwane, R., Melbourne, A., Doel, T., Dymarkowski, S., De~Coppi, P., et~al.: An automated framework for localization, segmentation and super-resolution reconstruction of fetal brain mri. NeuroImage  \textbf{206},  116324 (2020)

\bibitem{feurer-neurips15a}
Feurer, M., Klein, A., Eggensperger, K., Springenberg, J., Blum, M., Hutter, F.: Efficient and robust automated machine learning. In: Advances in Neural Information Processing Systems 28 (2015). pp. 2962--2970 (2015)

\bibitem{fox_mild_2018}
Fox, N.S., Monteagudo, A., Kuller, J.A., Craigo, S., Norton, M.E.: Mild fetal ventriculomegaly: diagnosis, evaluation, and management. American Journal of Obstetrics and Gynecology  \textbf{219}(1),  B2--B9 (Jul 2018). \doi{10.1016/j.ajog.2018.04.039}

\bibitem{kuklisova2012reconstruction}
Kuklisova-Murgasova, M., Quaghebeur, G., Rutherford, M.A., Hajnal, J.V., Schnabel, J.A.: Reconstruction of fetal brain mri with intensity matching and complete outlier removal. Medical image analysis  \textbf{16}(8),  1550--1564 (2012)

\bibitem{sanchez2024assessing}
Sanchez, T., Mihailov, A., Gomez, Y., Juan, G.M., Eixarch, E., Jakab, A., Dunet, V., Koob, M., Auzias, G., Cuadra, M.B.: Assessing data quality on fetal brain mri reconstruction: a multi-site and multi-rater study. In: International Workshop on Preterm, Perinatal and Paediatric Image Analysis. pp. 46--56. Springer (2024)

\bibitem{sanchez2024biometry}
Sanchez, T., Mihailov, A., Koob, M., Girard, N., Manchon, A., Valenzuela, I., G{\'o}mez-Chiari, M., Mart{\'\i}~Juan, G., Pron, A., Eixarch, E., et~al.: Biometry and volumetry in multi-centric fetal brain mri: assessing the bias of super-resolution reconstruction. medRxiv pp. 2024--09 (2024)

\bibitem{tourbier_automated_2017}
Tourbier, S., Velasco-Annis, C., Taimouri, V., Hagmann, P., Meuli, R., Warfield, S.K., Bach~Cuadra, M., Gholipour, A.: Automated template-based brain localization and extraction for fetal brain {MRI} reconstruction. NeuroImage  \textbf{155},  460--472 (2017). \doi{10.1016/j.neuroimage.2017.04.004}

\bibitem{uus2023retrospective}
Uus, A.U., Egloff~Collado, A., Roberts, T.A., Hajnal, J.V., Rutherford, M.A., Deprez, M.: Retrospective motion correction in foetal mri for clinical applications: existing methods, applications and integration into clinical practice. The British journal of radiology  \textbf{96}(1147),  20220071 (2023)

\bibitem{uus2023combined}
Uus, A.U., Hall, M., Payette, K., Hajnal, J.V., Deprez, M., Rutherford, M.A., Hutter, J., Story, L.: Combined quantitative t2* map and structural t2-weighted tissue-specific analysis for fetal brain mri: pilot automated pipeline. In: International Workshop on Preterm, Perinatal and Paediatric Image Analysis. pp. 28--38. Springer (2023)

\bibitem{uus2023bounti}
Uus, A.U., Kyriakopoulou, V., Makropoulos, A., Fukami-Gartner, A., Cromb, D., Davidson, A., Cordero-Grande, L., Price, A.N., Grigorescu, I., Williams, L.Z., et~al.: Bounti: Brain volumetry and automated parcellation for 3d fetal mri. bioRxiv  (2023)

\bibitem{uus2024scanner}
Uus, A.U., Neves~Silva, S., Aviles~Verdera, J., Payette, K., Hall, M., Colford, K., Luis, A., Sousa, H.S., Ning, Z., Roberts, T., et~al.: Scanner-based real-time 3d brain+ body slice-to-volume reconstruction for t2-weighted 0.55 t low field fetal mri. medRxiv pp. 2024--04 (2024)

\bibitem{10015091}
Xu, J., Moyer, D., Gagoski, B., Iglesias, J.E., Ellen~Grant, P., Golland, P., Adalsteinsson, E.: Nesvor: Implicit neural representation for slice-to-volume reconstruction in mri. IEEE Transactions on Medical Imaging pp.~1--1 (2023). \doi{10.1109/TMI.2023.3236216}

\bibitem{xu2023nesvor}
Xu, J., Moyer, D., Gagoski, B., Iglesias, J.E., Grant, P.E., Golland, P., Adalsteinsson, E.: Nesvor: implicit neural representation for slice-to-volume reconstruction in mri. IEEE transactions on medical imaging  \textbf{42}(6),  1707--1719 (2023)

\end{thebibliography}

\end{document}